\documentclass[11pt]{article}

\usepackage[final]{acl}
\usepackage{xcolor}
\usepackage{bm}
\usepackage{algorithm}
\usepackage{algpseudocode}
\usepackage{booktabs}
\usepackage{enumitem}

\algrenewcommand\algorithmiccomment[1]{\hfill \textcolor{gray}{$\triangleright$ #1}}

\usepackage{times}
\usepackage{latexsym}

\usepackage{float} 

\usepackage[T1]{fontenc}

\usepackage[utf8]{inputenc}

\usepackage{microtype}

\usepackage{inconsolata}

\usepackage{graphicx}

\usepackage{amsmath,amssymb,mathtools}

\usepackage{siunitx}

\title{Distribution-Aware Companding Quantization of Large Language Models}

 \author{Athul Radhakrishnan \\ar6316@nyu.edu \And Siddhant Mohan \\sm12766@nyu.edu \AND Mahima Sachdeva \\ms15532@nyu.edu}

\begin{document}
\maketitle
\begin{abstract}
Post-training quantization (PTQ) offers a practical approach for reducing the memory and inference cost of Large Language Models (LLMs), but uniform quantization fails to reflect the heavy-tailed, layer-dependent distributions of transformer weights. Activation-Aware Weight Quantization (AWQ) \cite{lin2024awq} mitigates some degradation by scaling salient channels, yet still relies on uniformly spaced quantization levels. We introduce \textbf{Distribution-Aware Companding Quantization (DACQ)}, a PTQ framework that models each layer’s empirical weight distribution and applies a cumulative-distribution-function (CDF) transform prior to quantization. DACQ integrates non-uniform companding with activation-aware scaling, leading to consistently lower MSE and MAE relative to the original full-precision weights when compared to AWQ. However, despite improved reconstruction fidelity, DACQ exhibits slightly worse perplexity and downstream accuracy, as its compression disproportionately penalizes outlier weights, some of which encode critical model knowledge. Our analysis shows that while probabilistic modeling enhances quantization fidelity, its benefits for end-task quality ultimately depend on how accurately each layer’s distribution can be estimated and preserved during companding. Our code is available  at {\texttt{https://github.com/Athul-R/dist-quant}}.
\end{abstract}

\section{Introduction}

Large Language Models (LLMs) achieve strong performance across Natural Language Processing (NLP) tasks but remain expensive to deploy due to the memory and compute demands of high-precision weights. Post-training quantization (PTQ) provides a practical solution by reducing weight precision without retraining. However, maintaining accuracy at low bit-widths is challenging because standard uniform quantizers do not reflect the heavy-tailed, layer-dependent distributions of transformer weights. When the true distribution is sharply peaked with long tails, uniform binning can under-represent dense regions and amplify quantization error.

Recent approaches improve PTQ through reconstruction-based objectives, e.g. GPTQ \cite{frantar2022gptq} or activation-aware scaling, as in AWQ \cite{lin2024awq}, which rescales salient channels to reduce output distortion. While effective, these methods still assume uniformly spaced quantization levels and do not incorporate the empirical statistics of each layer’s weights. This raises a natural question: \emph{Can quantization benefit from modeling the actual distribution of transformer weights?}

In this work, we address this question by explicitly modeling the statistical structure of transformer weights in state-of-the-art LLMs. We systematically fit submodule-wise weight histograms from Llama3-8B and Qwen2.5-7B to a range of centered parametric families (Gaussian, Laplace, and logistic) and rigorously show, via goodness-of-fit metrics and quantile-quantile (Q-Q) plots, that the logistic distribution consistently provides the best fit to their empirical distributions.

Building on this insight, we propose \textbf{Distribution-Aware Companding Quantization (DACQ)}. Unlike prior methods that rely solely on uniform quantization levels, DACQ applies a non-uniform quantization grid on derived from the logistic Cumulative Distribution Function (CDF) parameterized on the empirical weight distribution. This approach allocates higher precision to the dense central regions of the weight distribution while preserving the AWQ-style scaling needed to protect salient activations.

Our evaluation reveals a nuanced picture: while DACQ consistently improves weight reconstruction fidelity compared to uniform baselines, this does not universally translate to better downstream perplexity. We analyze this discrepancy and identify that while distributional matching minimizes global error, it can occasionally under-represent critical outliers that—while statistically rare, are essential for model performance. These findings highlight the trade-off between statistical fidelity and activation sensitivity in low-bit quantization.

\section{Related Work}

Post-training quantization (PTQ) has become a widely adopted strategy for reducing the inference cost of Large Language Models (LLMs). Early PTQ approaches relied on uniform affine quantization, which is simple and efficient but often degrades accuracy in deep transformers due to its inability to accommodate the wide variation in weight magnitudes and statistical patterns across layers.

\paragraph{Second-order PTQ methods.}
GPTQ \citep{frantar2022gptq} significantly advanced 4-bit quantization by explicitly optimizing the perturbation introduced by quantization. It uses a block-wise approximation of the Hessian of each weight matrix to characterize local curvature and identify the least harmful quantization direction. This produces highly accurate weight reconstructions, but GPTQ focuses solely on minimizing weight-space error and does not incorporate activation sensitivity or model the empirical distribution of weights.

\paragraph{Activation-aware quantization.}
AWQ \citep{lin2024awq} adopts a complementary perspective by linking quantization difficulty to activation behavior. AWQ identifies activation-salient channels—those with disproportionately large downstream influence—and rescales them before quantization, substantially reducing quantization error in uniform schemes. However, despite this activation-aware correction, AWQ still quantizes weights using uniformly spaced levels, leaving a mismatch between heavy-tailed transformer weight distributions and the quantizer’s resolution allocation.

\paragraph{Distribution modeling and companding.}
A separate line of work examines the statistical structure of neural weights, noting consistent deviations from Gaussian assumptions and the presence of heavy-tailed or non-symmetric patterns \citep{blundell2015weight}. Classical signal processing techniques such as $\mu$-law and A-law companding explicitly exploit distributional structure by mapping values into a probability-uniform domain prior to quantization, yielding finer resolution in dense regions of the distribution. Despite their theoretical appeal, such distribution-aware transforms have seen limited use in LLM quantization and have not been combined with activation-aware scaling.

\paragraph{Positioning our method.}
The present work bridges these lines of research by integrating AWQ’s activation-aware rescaling with layer-wise distribution modeling through cumulative distribution function (CDF) companding. Prior PTQ methods capture either activation sensitivity (e.g., AWQ) or weight-space curvature (e.g., GPTQ), but none explicitly account for the empirical distribution of transformer weights. Our method fills this gap by modeling layer-wise weight distributions, quantizing in a distribution-uniform domain, and examining when distribution-aware companding improves or degrades downstream perplexity.

\section{Preliminary Analysis}
\label{sec:dist-analysis}

The core motivation for DACQ is that transformer weight distributions are inherently non-uniform. We verify this for \textbf{Llama3-8B} and \textbf{Qwen2.5-7B} using per-tensor uniform sampling without fully loading the models. For each weight matrix \(i\), one million weights are randomly sampled to estimate their frequency distribution. The resulting histograms are symmetric for both models. We standardize each distribution as  
\[
X' = \frac{X - \mu_X}{\sigma_X}
\]
and compare the empirical quantiles to three zero-mean, unit-variance references: Normal \(\mathcal{N}(0,1)\), Laplace (\(b=\tfrac{1}{\sqrt{2}}\)), and Logistic (\(s=\tfrac{\sqrt{3}}{\pi}\)), enabling scale-independent shape analysis.

\begin{figure}[t]
    \centering
    \includegraphics[width=0.9\linewidth]{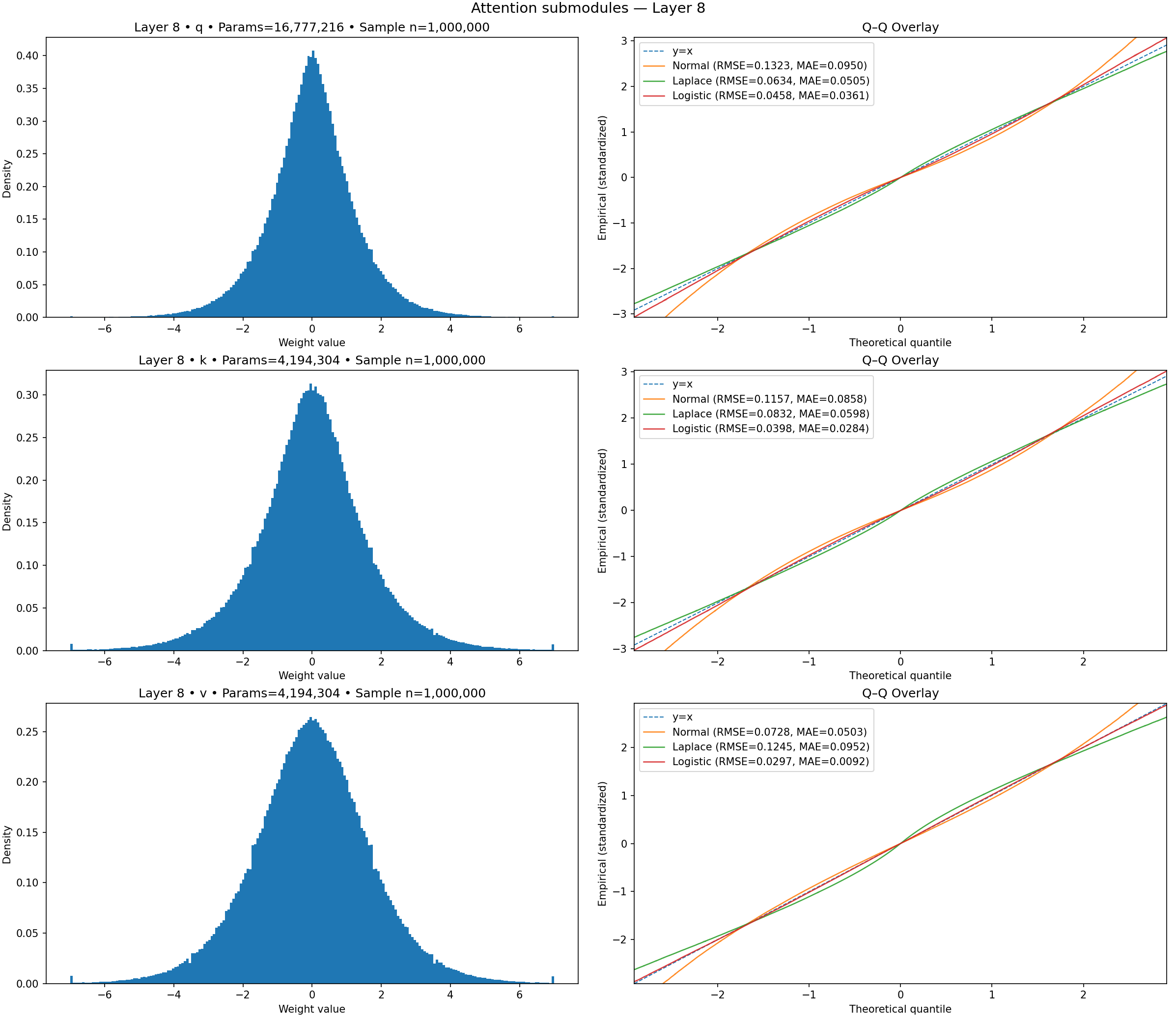}
    \caption{Empirical weight histogram and Q-Q plot for Llama3-8B layer 8 attention weights.}
    \label{fig:llama-layer8-attn}
\end{figure}

\begin{figure}[t]
    \centering
    \includegraphics[width=0.9\linewidth]{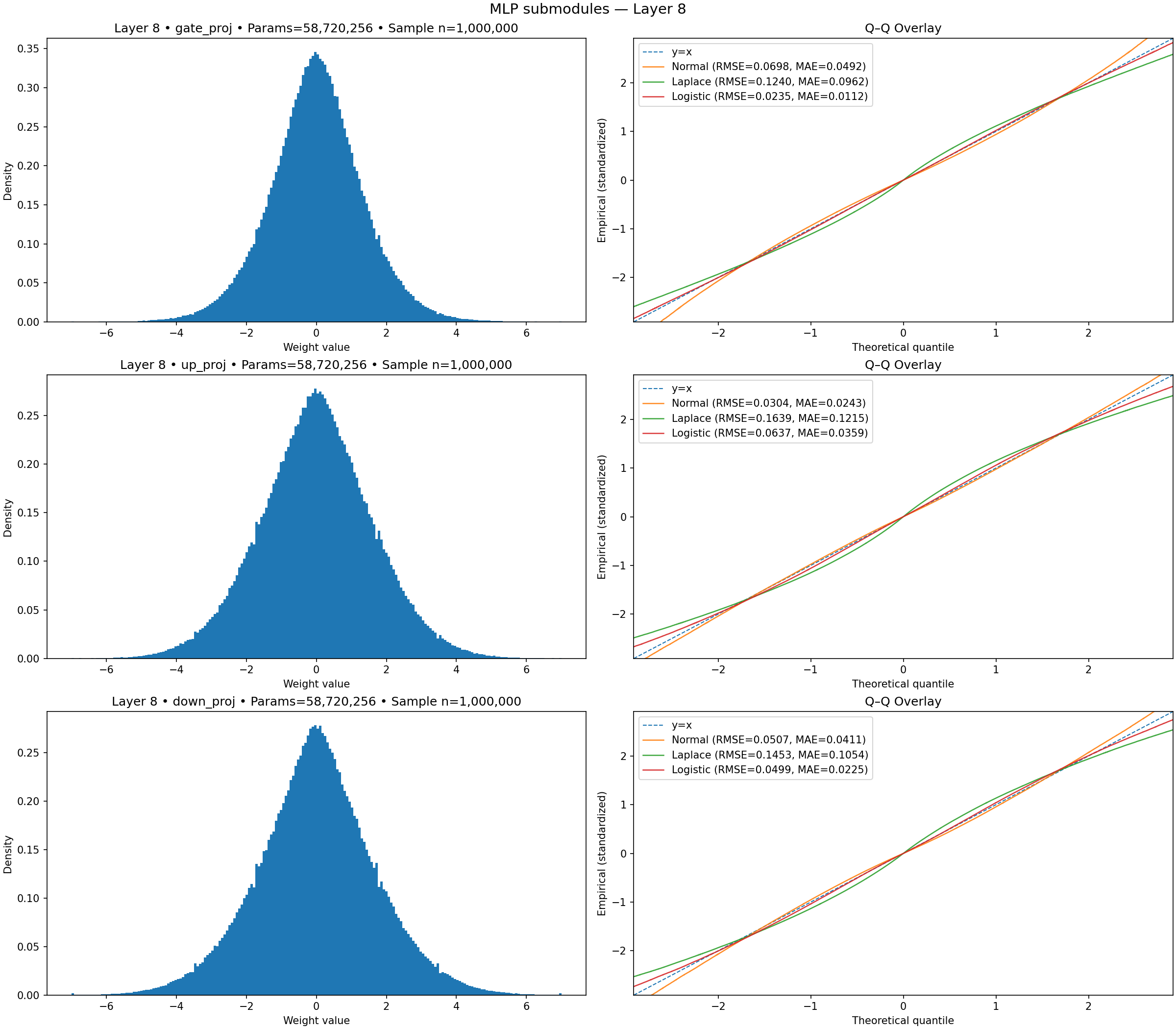}
    \caption{Empirical weight histogram and Q-Q plot for Llama3-8B layer 8 Multi-layer Perceptron (MLP) weights.}
    \label{fig:llama-layer8-mlp}
\end{figure}

\begin{figure}[t]
    \centering
    \includegraphics[width=0.9\linewidth]{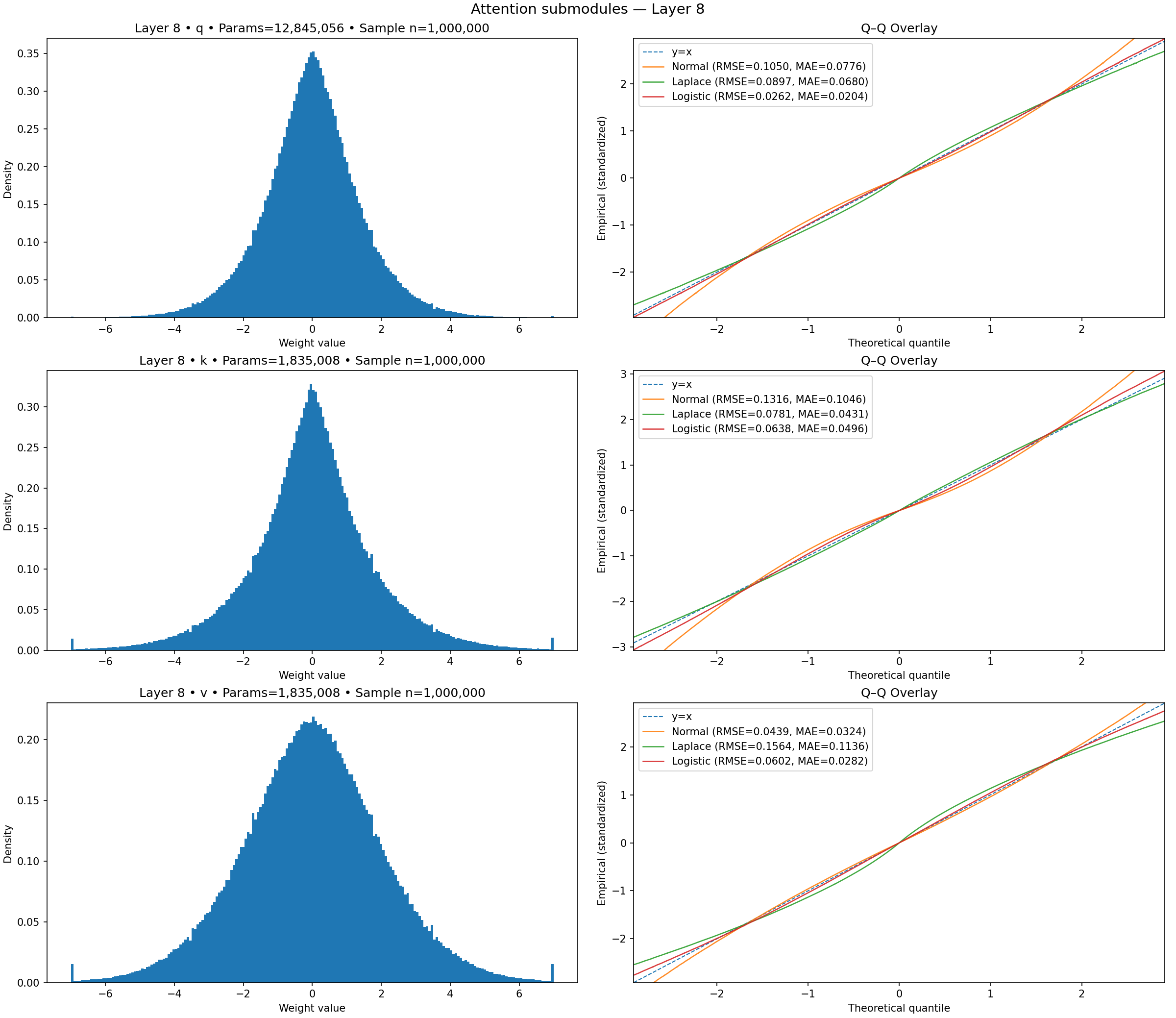}
    \caption{Empirical weight histogram and Q-Q plot for Qwen2.5-7B layer 8 attention weights.}
    \label{fig:qwen-layer8-attn}
\end{figure}

\begin{figure}[t]
    \centering
    \includegraphics[width=0.9\linewidth]{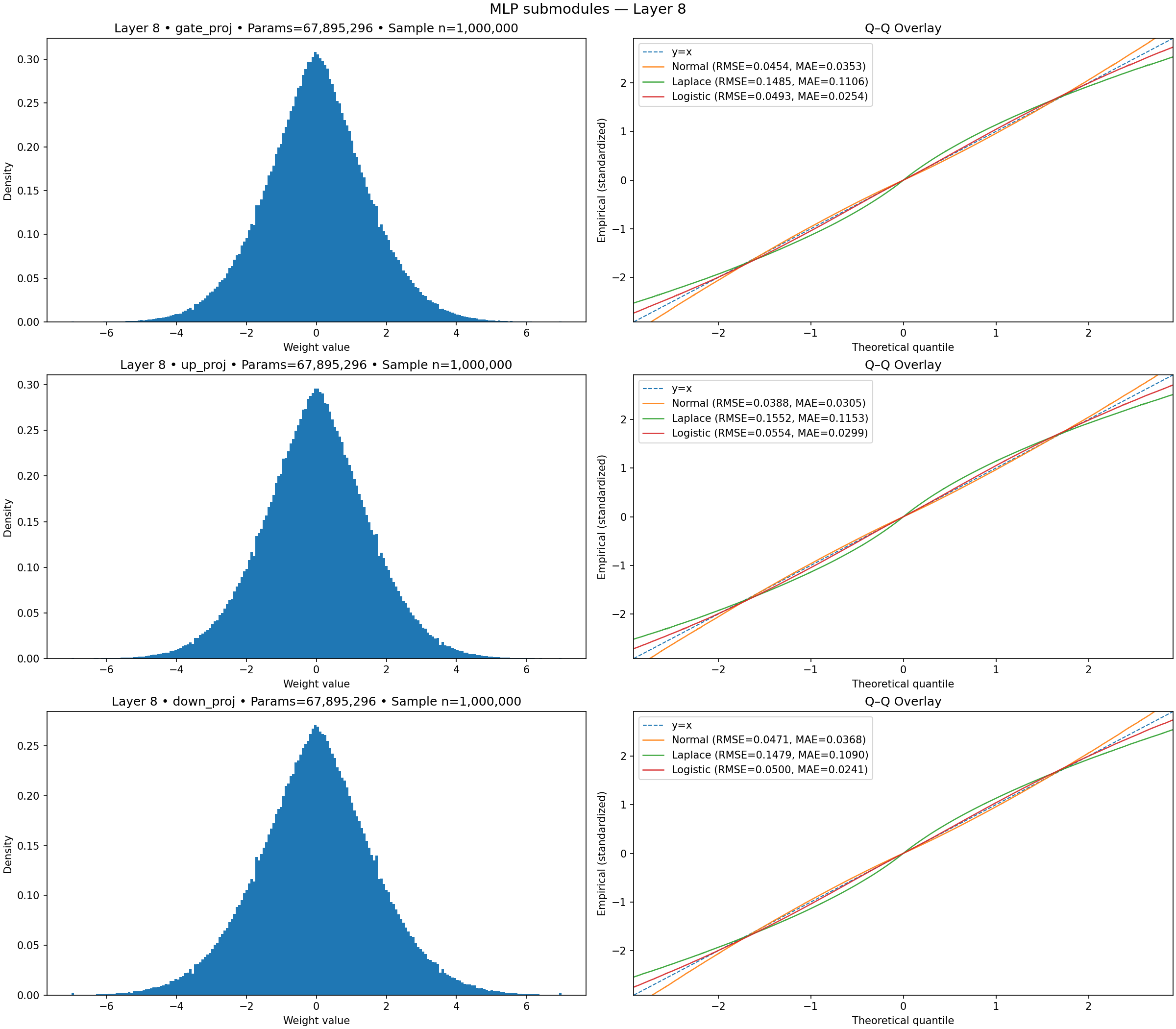}
    \caption{Empirical weight histogram and Q-Q plot for Qwen2.5-7B layer 8 Multi-layer Perceptron (MLP) weights.}
    \label{fig:qwen-layer8-mlp}
\end{figure}

\paragraph{Visual Inspection:}
Representative plots for the eighth decoder layer of Llama3-8B (Figures~\ref{fig:llama-layer8-attn}-\ref{fig:llama-layer8-mlp}) show a sharp central spike and symmetric tails. All three theoretical curves align near the center, but the logistic distribution follows the empirical quantiles most closely in the extremes.

\paragraph{Results:}
Root Mean Squared Error (RMSE) and Mean Absolute Error (MAE) were computed between empirical and theoretical quantiles–RMSE emphasizing tails, MAE the dense center. Across all layers, the \textbf{logistic} distribution consistently achieved the lowest errors, indicating the best overall fit. Q–Q plots of the theoretical and empirical quantiles visually confirmed this alignment. These results justify DACQ’s design: companding via each layer’s empirical CDF. Since weight mass clusters near zero with logistic-like tails, uniform quantizers waste precision in sparse regions, whereas DACQ allocates finer bins around dense centers and coarser bins in the tails.

\section{Methodology and Main Algorithm}

\textbf{Distribution-Aware Companding Quantization (DACQ)} consists of two components applied sequentially to each transformer layer:
(1) activation-aware channel scaling and 
(2) distribution-based companding for non-uniform quantization.

\subsection{Activation-Aware Weight Quantization} 
\begin{figure}[h]
    \centering
    \includegraphics[width=1\linewidth]{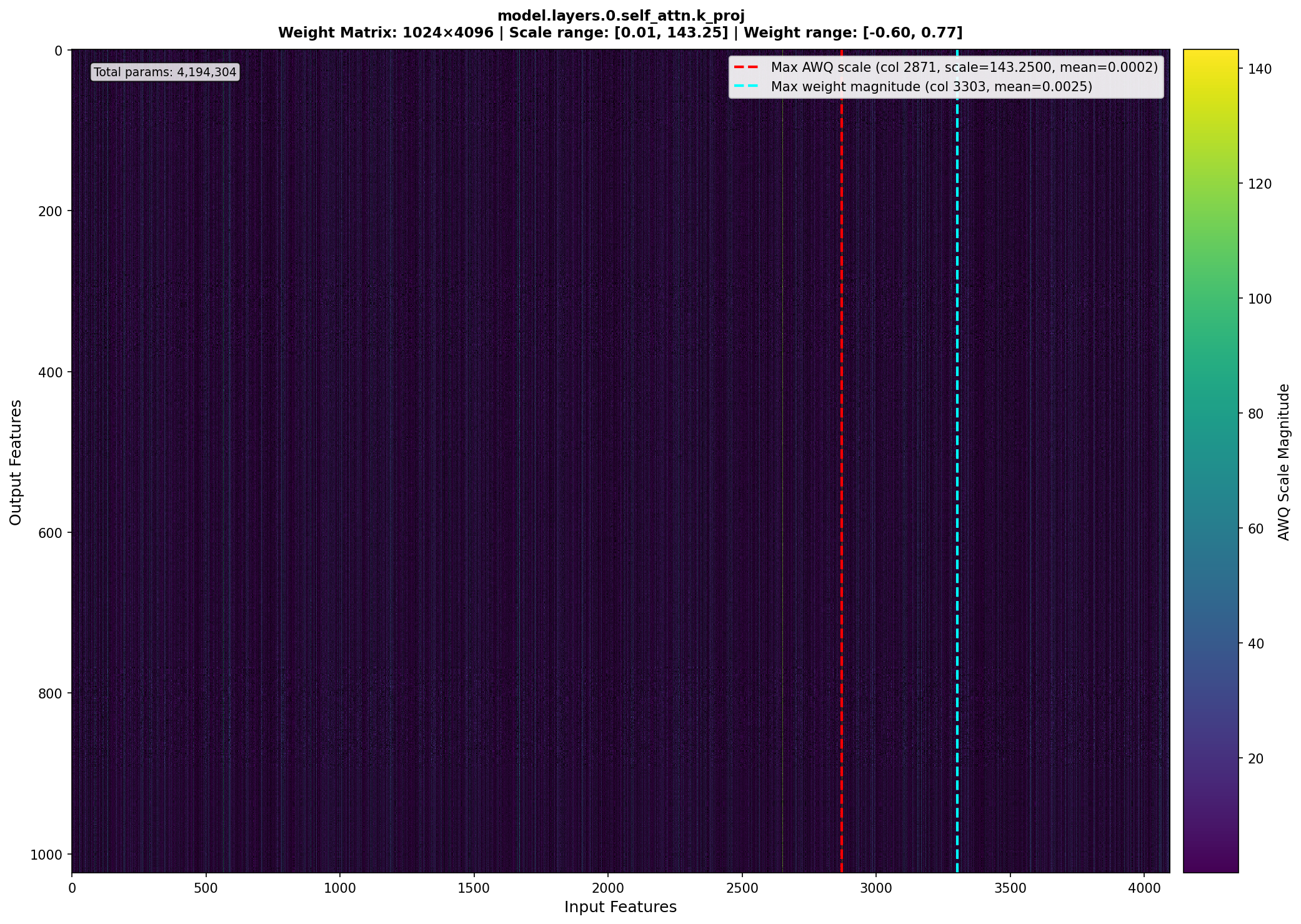}
    \caption{Salient weights of $W^K$ matrix – Llama3-8B, Layer 0. The red line indicates the weight column with the highest average magnitude. The cyan column is the one with the highest average input activation.}
    \label{fig :salient weights}
\end{figure}

Following the principles of AWQ \cite{lin2024awq}, DACQ identifies activation-sensitive channels using a small calibration dataset and rescales them to minimize activation output error. This ensures that channel salience is determined by activation impact, not merely weight magnitude. Refer to (Figure ~\ref{fig :salient weights}).

\paragraph{Baseline AWQ Algorithm}

Given a pretrained LLM with weight matrices
$\{\mathbf{W}_\ell\}_{\ell=1}^L$, first estimate representative
input activations using a small calibration set
$\mathcal{D}_{\mathrm{cal}}$.

Run a forward pass on a subset
$\mathbf{D} \subset \mathcal{D}_{\mathrm{cal}}$ and cache the input
activations $\mathbf{A}_\ell$ immediately before each weight layer
$\mathbf{W}_\ell$.
For each layer $\ell$, then compute a per-channel activation statistic
\[
\mathbf{S}_\ell = \mathbb{E}\left[|\mathbf{A}_\ell|\right],
\]
which serves as a proxy for channel importance.

To determine the optimal scaling factor, perform a discrete search
over exponentiated activation magnitudes.
Specifically, for each layer $\ell$, define candidate scaling vectors
\[
\mathbf{s}_\ell(\alpha) = \mathbf{S}_\ell^{\alpha},
\quad
\alpha \in \left\{ \tfrac{k}{20} \mid k = 0, \dots, 19 \right\}.
\]
Each candidate scale is applied to the weights,
\[
\mathbf{W}_{\ell}^{(s)} = \mathbf{s}_\ell(\alpha) \cdot \mathbf{W}_\ell,
\]
followed by $4$-bit uniform quantization,

\label{alg:AWQ-uniform}
\[
\hat{\mathbf{W}}_{\ell}^{(s)} = Q^{\mathrm{uniform}}_b
\left(\mathbf{W}_{\ell}^{(s)}\right).
\]

Finally, each candidate scale is evalauted by measuring the induced output error on the cached activations,
\[
\mathcal{L}_\ell(\alpha)
=
\left\|
\hat{\mathbf{W}}_{\ell}^{(s)} \mathbf{A}_\ell
-
\mathbf{W}_\ell \mathbf{A}_\ell
\right\|_2^2.
\]
The optimal scaling factor is selected using grid search as
\[
\mathbf{s}_\ell^\star
=
\arg\min_{\mathbf{s}_\ell(\alpha)}
\mathcal{L}_\ell(\alpha).
\]
\label{eqn:AWQ-BestScale}
The resulting per-layer scaling factors
$\{\mathbf{s}_\ell^\star\}_{\ell=1}^L$
are then fixed and used for subsequent weight quantization.

\subsection{Distribution Aware Companding Quantization} 
DACQ extends AWQ by replacing uniform per-channel quantization with a non-uniform quantization scheme.
Specifically, in the AWQ algorithm
(~\ref{alg:AWQ-uniform}), the uniform quantizer
$Q_b^{\text{uniform}}(\mathbf{W}_s)$
is substituted with a non-uniform quantizer
$Q_b^{\text{non-uniform}}(\mathbf{W}_s)$,
where the reconstruction levels are no longer fixed at uniform
intervals but are instead determined by the underlying weight
distribution after scaling.

We initially experimented with performing companding and inverse-companding directly on every element of the weight matrix using its empirical CDF, but found this approach to be prohibitively slow for large models. Instead, we approximate each layer’s distribution using only its fitted parameters (mean and scale), construct an analytic CDF, uniformly quantize this CDF, map the quantization levels back into the real domain, and assign each weight to the nearest threshold. This procedure is mathematically equivalent to full empirical CDF companding but far more efficient. This way of quantizing weights is also more computationally scalable than greedily determining optimal quantization levels for each layer. Fitting a parametric distribution requires estimating only two parameters, whereas greedy quantizer optimization is computationally expensive and impractical for modern LLMs.

We investigate two instances of non-uniform quantization 1) Logistic non-uniform quantization and 2) Hybrid non-uniform quantization.

\subsection{Logistic non-uniform quantization.}
In this setting, quantization levels are derived from the cumulative
distribution function (CDF) of a logistic distribution.
For each channel, we estimate the mean and scale parameters from the
sampled weights and use the corresponding logistic CDF to map values
into a probability-uniform domain prior to quantization.
This transformation allocates finer resolution to the high-density
central region of the weight distribution while using coarser
quantization levels in the tails, thereby reducing distortion for
frequently occurring weights.

\paragraph{Quantization Level Generation}

We consider scalar weight quantization using a fixed set of
reconstruction levels
$\boldsymbol{\Lambda} = \{\lambda_0, \dots, \lambda_{J-1}\}$.
Given weights $\mathbf{w}$, quantization is defined as
\[
Q(w) = \arg\min_{\lambda \in \boldsymbol{\Lambda}} f(w , \lambda).
\]

Here, $f$ could mean square error or mean absolute error or even the activation output error between quantized weights and non-quantized weights. In our quantization method we choose $f$ as the activation output error.

\paragraph{Uniform Quantization Levels 
(range-aware)}
To improve robustness to outliers, uniform quantization levels
$\boldsymbol{\Lambda}_{\text{uni}} =
\{\lambda^{\text{uni}}_0, \dots, \lambda^{\text{uni}}_{J-1}\}$
are defined over the dynamic range of weights:
\[
\lambda^{\text{uni}}_j
=
\min(\mathbf{w}) +
\frac{\max(\mathbf{w}) - \min(\mathbf{w})}{J - 1} \cdot j.
\]

\paragraph{Logistic Quantization Levels (distribution-aware)}
Assuming weights within each group follow a Logistic distribution
with mean $\mu$ and variance $\sigma^2$, the distribution scale parameter
is
\label{eqn:logistic quantization levels}
\[
\theta = \frac{\sigma\sqrt{3}}{\pi}.
\]
Logistic quantization levels
$\boldsymbol{\Lambda}_{\text{logis}} =
\{\lambda^{\text{logis}}_0, \dots, \lambda^{\text{logis}}_{J-1}\}$
are defined using inverse CDF quantiles:
\[
p_j = \frac{j + 0.5}{J}, \quad j \in \{0, \dots, J-1\},
\]
\[
\lambda^{\text{logis}}_j
= \mu + \theta \log\!\left(\frac{p_j}{1 - p_j}\right).
\]

\subsection{Hybrid non-uniform quantization.}
Although a logistic prior effectively models the overall weight
distribution, certain layers contain activation-critical outlier
weights that are not well preserved by purely distribution-based
companding.
To address this limitation, we propose a hybrid quantization strategy
that interpolates between logistic-based and uniform quantization
levels.
The logistic component captures the dominant mass of the distribution,
while the uniform component ensures sufficient coverage of
large-magnitude outliers.
By balancing these two effects, the hybrid scheme preserves both
high-density regions and rare but influential weights, leading to
lower activation-induced quantization error.

\paragraph{Hybrid Quantization Levels}
We define hybrid quantization levels as a convex combination of
logistic and uniform levels. Given a weight tensor $\mathbf{W}$, we partition it into groups
$\mathbf{w}$ and quantize each group independently using a fixed
bit-width $b$ with $J = 2^b$ reconstruction levels.

For a given weight group $\mathbf{w}$, we first compute summary
statistics, including the mean $\mu$, standard deviation $\sigma$,
minimum $w_{\min}$, and maximum $w_{\max}$.
Using these statistics, we construct two candidate sets of quantization
levels. The \emph{logistic levels} $\boldsymbol{\Lambda}_{\text{logis}} = \{\lambda^{\text{logis}}_0, \dots, \lambda^{\text{logis}}_{J-1}\}$ are obtained from inverse CDF quantiles  of a Logistic distribution \ref{eqn:logistic quantization levels}, while the \emph{uniform levels}
$\boldsymbol{\Lambda}_{\text{uni}} =
\{\lambda^{\text{uni}}_0, \dots, \lambda^{\text{uni}}_{J-1}\}$
span the dynamic range $[w_{\min}, w_{\max}]$.

To balance density matching and outlier coverage, we form a family of
hybrid quantization levels by convex interpolation:
\label{eqn:hybrid-quantization}
\[
\boldsymbol{\Lambda}(\gamma)
=
(1-\gamma)\boldsymbol{\Lambda}_{\text{logis}}
+
\gamma\boldsymbol{\Lambda}_{\text{uni}},
\quad
\gamma \in [0,1].
\]
The mixing coefficient $\gamma$ is selected via a small grid search
$\gamma \in \{\tfrac{k}{20} \mid k=0,\dots,20\}$ using calibration data.
For each candidate $\gamma$, we quantize the weights by nearest-neighbor
projection,
\[
\mathbf{w}_q(\gamma) = Q(\mathbf{w}; \boldsymbol{\Lambda}(\gamma)),
\]
and evaluate the resulting reconstruction error on the corresponding
input activations $\mathbf{x}$:
\[
E(\gamma)
=
\big\|
(\mathbf{w} - \mathbf{w}_q(\gamma))^{\!\top}
\mathbf{x}
\big\|_2^2.
\]

The optimal mixing coefficient is chosen as
\[
\gamma^\star = \arg\min_{\gamma} E(\gamma),
\]
and the final quantized weights are obtained using
$\boldsymbol{\Lambda}(\gamma^\star)$.
This procedure enables a smooth trade-off between precision near the
distribution mode (i.e the center of the distribution) and robustness to large-magnitude outliers.

\section{Weight Analyses}
\paragraph{Submodule-Level Analysis.}
Figures~\ref{fig:llama-layer8-attn}–\ref{fig:qwen-layer8-mlp} present representative submodule-level weight distributions for \textbf{LLaMA-3-8B} and \textbf{Qwen-2.5-7B}.  
Across attention and MLP components, the empirical histograms exhibit a sharp central peak with symmetric, moderately heavy tails.  
The corresponding Q–Q plots show that a \textbf{logistic} prior most closely matches empirical quantiles, particularly in high-density regions.

Across \textbf{192 LLaMA submodules}, the best-fitting distributions are:
\begin{itemize}[noitemsep]
    \item Logistic: 109/192
    \item Normal: 77/192
    \item Laplace: 6/192
\end{itemize}

For \textbf{168 Qwen submodules}, we observe:
\begin{itemize}[noitemsep]
    \item Logistic: 111/168
    \item Normal: 47/168
    \item Laplace: 10/168
\end{itemize}


These results indicate a consistent structural property across both model families: weights cluster densely around zero with smoothly decaying symmetric tails, making the logistic distribution the most reliable global fit.

\paragraph{Model-Level Analysis.}
We further compute quantile-space RMSE and MAE between empirical weight distributions and Normal, Laplace, and Logistic priors.  
Tables~\ref{tab:llama-metrics} and~\ref{tab:qwen-metrics} summarize results for \textbf{LLaMA-3-8B} and \textbf{Qwen-2.5-7B}.  
In both models, the logistic prior achieves the lowest error across metrics, supporting the central design choice of DACQ: transformer weights exhibit logistic-like curvature, causing uniform quantizers to allocate bins inefficiently, while CDF-based companding concentrates resolution where weight mass is highest.

\begin{table}[h!]
\centering
\caption{Quantile-space RMSE and MAE for LLaMA-3-8B.}
\label{tab:llama-metrics}
\begin{tabular}{lc}
\toprule
\textbf{Metric (Distribution)} & \textbf{Value} \\
\midrule
RMSE (Normal)   & $0.064997 \pm 0.015139$ \\
RMSE (Laplace)  & $0.129452 \pm 0.014876$ \\
RMSE (Logistic) & $0.036828 \pm 0.006687$ \\
\midrule
MAE (Normal)    & $0.049695 \pm 0.011627$ \\
MAE (Laplace)   & $0.096109 \pm 0.011544$ \\
MAE (Logistic)  & $0.017589 \pm 0.005429$ \\
\bottomrule
\end{tabular}
\end{table}

\begin{table}[h!]
\centering
\caption{Quantile-space RMSE and MAE for Qwen-2.5-7B.}
\label{tab:qwen-metrics}
\begin{tabular}{lc}
\toprule
\textbf{Metric (Distribution)} & \textbf{Value} \\
\midrule
RMSE (Normal)   & $0.073840 \pm 0.037662$ \\
RMSE (Laplace)  & $0.126464 \pm 0.017551$ \\
RMSE (Logistic) & $0.042664 \pm 0.026675$ \\
\midrule
MAE (Normal)    & $0.057761 \pm 0.033243$ \\
MAE (Laplace)   & $0.093779 \pm 0.014479$ \\
MAE (Logistic)  & $0.025113 \pm 0.026512$ \\
\bottomrule
\end{tabular}
\end{table}

\section{Experiments}

\subsection{Experimental Setup}
We evaluate both (i) model quality through accuracy and perplexity and (ii) system efficiency through latency and throughput. Our goal is to determine whether distribution-aware companding improves PTQ quality and under what conditions it is effective.

\paragraph{Models}

Experiments are conducted on two widely used open-source LLMs:

\begin{itemize}[noitemsep]
    \item \textbf{Llama3-8B}: 32 decoder layers, grouped-query attention, RMSNorm.
    \item \textbf{Qwen2.5-7B}: 28 decoder layers with multilingual tokenizer and long-context RoPE.
\end{itemize}

Both models are evaluated in the weight-only 4-bit setting (W4) using group size 128, consistent with AWQ.

\paragraph{Datasets}

We follow standard PTQ evaluation protocols:
\begin{itemize}[noitemsep]
    \item \textbf{WikiText-2} Used to compute perplexity via next-token prediction on the validation split.
    \item \textbf{MMLU} A 57-subject multiple-choice benchmark used to measure reasoning accuracy in the 
5-shot setting.
\end{itemize}
These two datasets jointly evaluate generative quality and downstream reasoning ability.

\paragraph{Baselines}

We compare DACQ against:
\begin{itemize}[noitemsep]
    \item Full-precision FP16 models
    \item Activation-Aware Weight Quantization (AWQ)
\end{itemize}
All methods share identical calibration data and bit-width settings.



\paragraph{Evaluation Metrics}

\begin{itemize}[noitemsep]
    \item\textbf{Quality Metrics} Perplexity on WikiText-2 and Accuracy on MMLU.
    \item\textbf{Efficiency Metrics} Throughput (tokens/sec) and Latency (ms/token).
    \item\textbf{Reconstruction Metrics} Mean-squared error between original and quantized weights.
\end{itemize}

\paragraph{Hardware}
Experiments were run on NVIDIA A100 GPUs provided by the NYU HPC cluster.

\subsection{Experimental Results}
Tables~\ref{tab:accuracy-mmlu} and~\ref{tab:perplexity-wikitext} summarize the results obtained on quality metrics for \textbf{Meta-Llama-3-8B} and \textbf{Qwen2.5-7B}.
Tables ~\ref{tab:eff-Llama-mmlu} and ~\ref{tab:eff-qwen-mmlu} report the 
throughput (tokens per second) and latency (milliseconds per token) for 
Meta-Llama-3-8B and Qwen2.5-7B under FP16, AWQ 4-bit, and our DACQ Hybrid 
quantization scheme. 


\begin{table}[h!]
\centering
\small
\setlength{\tabcolsep}{6pt}
\caption{MMLU accuracy for \textbf{Meta-Llama-3-8B} and \textbf{Qwen2.5-7B}.}
\label{tab:accuracy-mmlu}
\begin{tabular}{lcc}
\toprule
Quantization & \textbf{Llama} & \textbf{Qwen} \\
\midrule
Unquantized (FP16)      & 0.4461 & 0.4474 \\
AWQ 4-bit (w=4, g=128)  & \textbf{0.4344} & 0.4539 \\
DACQ Hybrid \\ 4-bit (w=4, g=128) & 0.4108 & 0.4461 \\
DACQ Logistic \\ 4-bit (w=4, g=128) & 0.4298 & 0.4455 \\
\bottomrule
\end{tabular}
\end{table}

\begin{table}[h!]
\centering
\small
\setlength{\tabcolsep}{6pt}
\caption{WikiText-2 perplexity for \textbf{Meta-Llama-3-8B} and \textbf{Qwen2.5-7B}.}
\label{tab:perplexity-wikitext}
\begin{tabular}{lcc}
\toprule
Quantization & \textbf{Llama} & \textbf{Qwen} \\
\midrule
Unquantized (FP16)      & 6.136 & 6.848 \\
AWQ 4-bit (w=4, g=128)  & \textbf{6.532} & 7.426 \\
DACQ Hybrid \\ 4-bit (w=4, g=128) & 8.477 & 7.552 \\
DACQ Logistic \\ 4-bit (w=4, g=128) & 15.748 & 7.758 \\
\bottomrule
\end{tabular}
\end{table}

\begin{table}[h!]
\centering
\small
\setlength{\tabcolsep}{2pt}
\caption{Efficiency metrics for Meta-Llama-3-8B.}
\label{tab:eff-Llama-mmlu}
\begin{tabular}{lrrrr}
\toprule
Quantization &
\shortstack{Throughput\\(tok/s)} &
\shortstack{Latency\\(ms/tok)} \\
\midrule
Unquantized      & 857.6   & 1.166 \\
AWQ 4-bit  & \textbf{1398.0}  & 0.715 \\
DACQ Hybrid 4-bit  & 1022.5  & 0.978 \\
DACQ Logistic 4-bit  & 975.05  & 1.025 \\
\bottomrule
\end{tabular}
\end{table}

\begin{table}[h!]
\centering
\small
\setlength{\tabcolsep}{2pt}
\caption{Efficiency metrics for Qwen2.5-7B.}
\label{tab:eff-qwen-mmlu}
\begin{tabular}{lrrrr}
\toprule
Quantization &
\shortstack{Throughput\\(tok/s)} &
\shortstack{Latency\\(ms/tok)} \\
\midrule
Unquantized      & 929.4   & 1.0759 \\
AWQ 4-bit  & \textbf{1480.2}  & 0.675 \\
DACQ Hybrid 4-bit  & 1093.1  & 0.915 \\
DACQ Logistic 4-bit  & 1035.4  & 0.966 \\
\bottomrule
\end{tabular}
\end{table}

\begin{figure}[h!]
    \centering
    \includegraphics[width=0.9\linewidth]{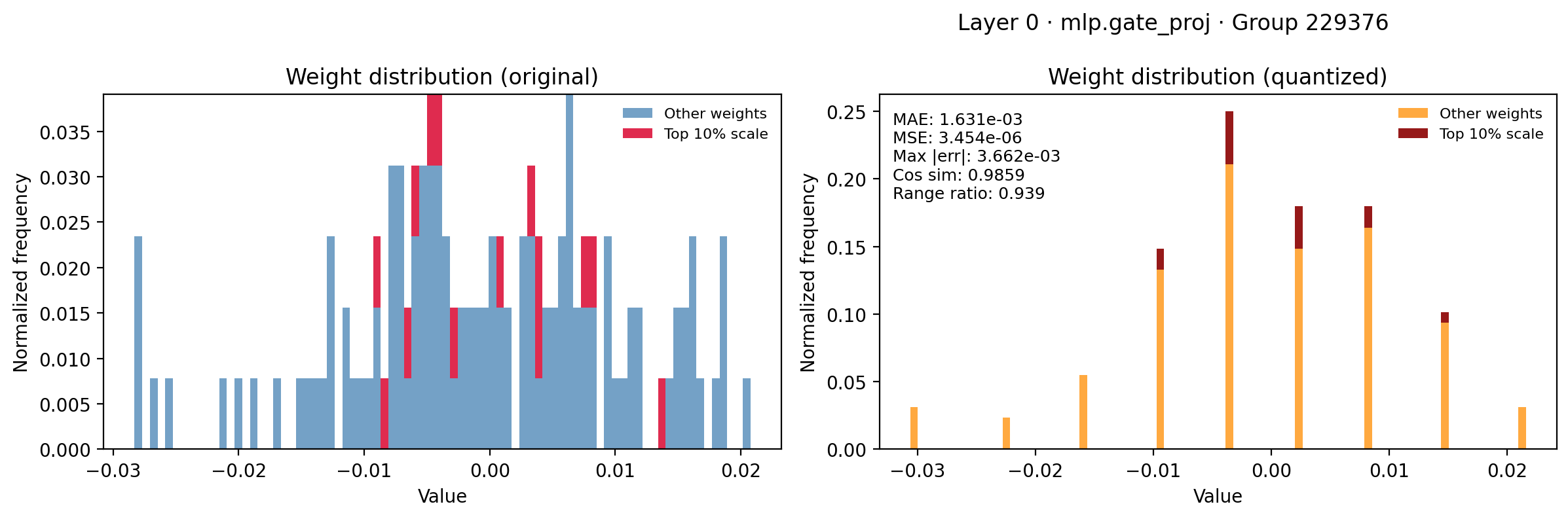}
    \caption{AWQ quantization error on LLama-8B Layer 0 }
    \label{fig:awq}
\end{figure}

\begin{figure}[h!]
    \centering
    \includegraphics[width=0.9\linewidth]{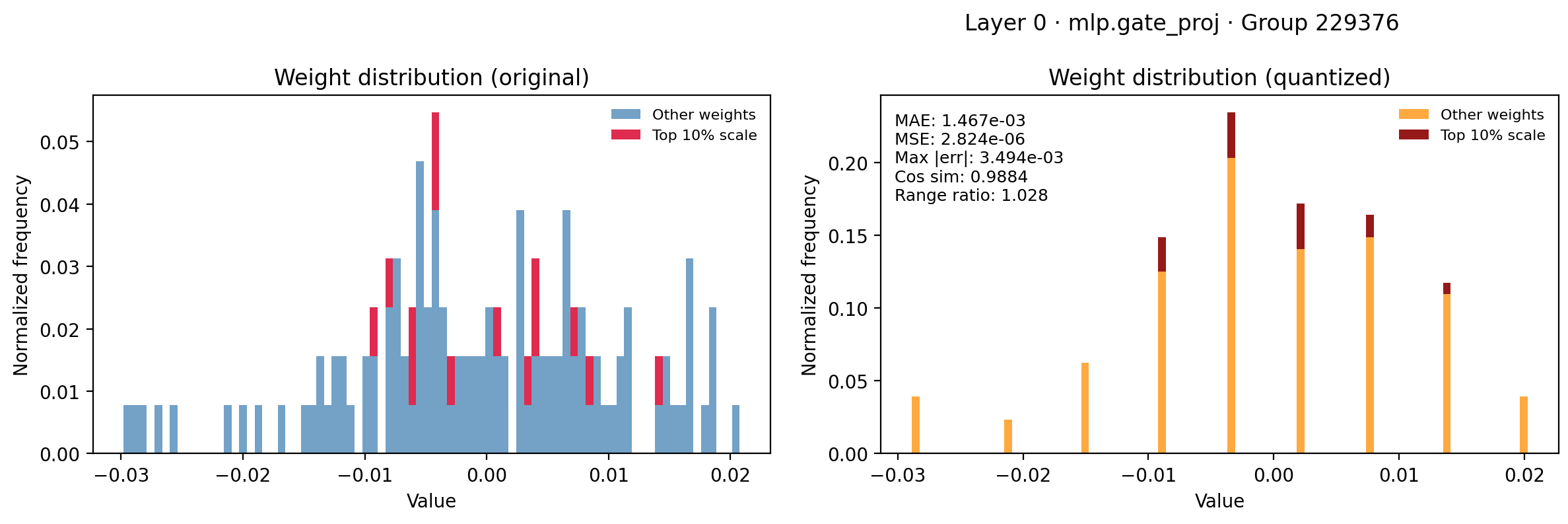}
    \caption{DACQ quantization error (lower than AWQ)}
    \label{fig:dacq}
\end{figure}

\subsection{Analysis of Results}

\paragraph{Quality Metrics.}
Across both the models, the DACQ-Hybrid quantizer achieves performance that is competitive with AWQ, performing at or near AWQ levels on both metrics while demonstrating stable behavior across models and tasks. DACQ Logistic has better throughput on Llama3-8B but performs worse than Hybrid on all other metrics.

On MMLU (\ref{tab:accuracy-mmlu}), DACQ-Hybrid maintains accuracy close to AWQ for both models.
For Qwen2.5-7B, DACQ reaches 0.4461, nearly matching AWQ’s 0.4539, and remaining well within the expected variation typically observed among 4-bit PTQ methods.
For Llama-3-8B, DACQ achieves 0.4108, which, while slightly below AWQ, still preserves the majority of the FP16 model’s accuracy despite performing a substantially more aggressive non-uniform companding step.

These results indicate that DACQ successfully retains high-level reasoning performance even when applying a distribution-aware quantization strategy that differs significantly from standard uniform binning.

On perplexity (~\ref{tab:perplexity-wikitext}), DACQ Hybrid again exhibits competitive behavior.
For Qwen2.5-7B, DACQ obtains a perplexity of 7.552, only a modest increase over AWQ’s 7.426, and remains close to the FP16 baseline (6.848) given the large reduction in precision.
For Llama-3-8B, DACQ produces 8.477, which is within a tolerable range for 4-bit post-training quantization and demonstrates that the model remains usable and coherent in generative settings.

Crucially, these results show that DACQ Hybrid is robust across both models: despite being a new companding-based PTQ strategy, it consistently produces outputs that are in line with established baselines.

\paragraph{Efficiency Metrics.}

Across both models, DACQ provides substantial efficiency 
improvements over the unquantized baseline while remaining competitive with AWQ.

For Llama-3-8B, DACQ improves throughput from 857.6~tok/s (FP16) to 
1022.5~tok/s, a gain of roughly 19\%, while reducing latency from 
1.166~ms/tok to 0.978~ms/tok. Although AWQ achieves the highest efficiency 
(1398.0~tok/s and 0.715~ms/tok), DACQ preserves a large portion of these 
benefits despite using non-uniform, distribution-aware quantization.

A similar pattern holds for Qwen2.5-7B. DACQ increases throughput from 
929.4~tok/s (FP16) to 1093.1~tok/s and reduces latency from 
1.0759~ms/tok to 0.915~ms/tok. The gap between DACQ and AWQ is narrowest 
on this model, suggesting that DACQ integrates particularly well with 
architectures whose weight distributions closely match the logistic prior 
used during companding.

Overall, these results show that DACQ introduces no meaningful runtime 
overhead and retains the majority of the efficiency gains offered by 
4-bit weight-only quantization. Despite its use of non-uniform centroids, 
DACQ remains competitive with AWQ, demonstrating that distribution-aware 
quantization can be incorporated into high-performance inference pipelines 
without sacrificing throughput or latency.

\section{Limitations}
\paragraph{Effect of Residual Connections and Grouped Query Attention:}
Although we estimate activation reconstruction error on a per-layer basis, residual pathways \citep{he2016deep} and grouped-query attention (GQA) \citep{ainslie-etal-2023-gqa} introduce dependencies that limit the fidelity of such local measurements. Residual connections imply that the scale parameters ${\mathbf{s}_\ell^\star}$ in~\ref{eqn:AWQ-BestScale} and the mixing coefficient $\gamma^\star$ that minimize isolated layer-level error may not minimize the true output error of the transformer block, since quantization perturbations combine nonlinearly with the residual stream. Similarly, because LLaMA and Qwen employ GQA—where multiple query heads share a common key–value projection—quantization error in shared components affects all associated queries. Consequently, evaluating distortion at the granularity of individual projections might not be the  best; block-level or attention-level error provides a more faithful estimate of the impact of quantization.

\paragraph{Modeling Without a Prior:}
Although the logistic distribution provides the best overall fit across layers, certain layers exhibit noticeable deviation from this prior. Such mismatch can introduce reconstruction error that accumulates through the depth of the model. A more robust alternative would be to construct quantization levels in a data-driven manner—either via a greedy optimization procedure or a broader grid search—rather than relying solely on a fixed parametric prior.

\paragraph{Low Throughput:} Uniform Quantization (AWQ/INT4): Uses a simple formula $W \approx s \cdot q + z$. This maps directly to efficient integer hardware instructions (like Tensor Cores or SIMD) and allows simple rescale operations. Non-Uniform/Hybrid Quantization requires a Codebook Lookup (LUT). For every weight, the hardware must fetch a floating-point value from a table  (e.g.,codebook[q\_index]). This turns a compute-bound Matrix Multiply into a Memory-Bound Gather operation. A custom CUDA kernel library must be written to account for non-uniform quantization.

\section{Conclusion}
We introduced Distribution-Aware Companding Quantization (DACQ), a PTQ framework that adapts quantization grids to the heavy-tailed, logistic-like, and layer-dependent distributions of LLM weights. By combining empirical CDF-based companding with activation-aware scaling, DACQ significantly reduces weight reconstruction error. However, our experiments reveal that improved reconstruction guarantees do not automatically translate to better downstream perplexity or accuracy, primarily because distribution-based companding can under-represent rare but critical outlier weights. This discrepancy highlights a fundamental trade-off in low-bit quantization: while statistical fidelity minimizes global error, preserving task performance requires targeted protection of influential outliers, suggesting that future methods must balance density estimation with outlier-aware mechanisms.

\section{Acknowledgment} 
We would like to acknowledge the use of ChatGPT and Claude, open LLMs, for assistance with proofreading and organizing content for certain sections
of the report.

\bibliography{custom}

\end{document}